# Algorithm For 3D-Chemotaxis Using Spiking Neural Network

Jayesh Choudhary[1], Vivek Saraswat[1], and Udayan Ganguly[1]

Department of Electrical Engineering, Indian Institute Of Technology Bombay, Mumbai, India

**Abstract.** In this work, we aim to devise an end-to-end spiking implementation for contour tracking in 3D media inspired by chemotaxis, where the worm reaches the region which has the given set concentration. For a planer medium, efficient contour tracking algorithms have already been devised, but a new degree of freedom has quite a few challenges. Here we devise an algorithm based on klinokinesis - where the worm's motion is in response to the stimuli but not proportional to it. Thus the path followed is not the shortest, but we can track the set concentration successfully. We are using simple LIF neurons for the neural network implementation, considering the feasibility of its implementation in the neuromorphic computing hardware.

## 1 Introduction and Motivation

Spiking Neural Networks are artificial neural networks that have more resemblance to the biological neural network. Here, the information is transferred from one neuron to another via discrete events, i.e., binary spikes allowing the network to relay information faster and in an energy-efficient manner. There is hence a significant interest in developing SNN based control applications for navigation and robotics.

Most of the neurobiological studies are performed on the worm C. Elegans because of its simple neural network. Despite being a simple organism, many of the molecular signals controlling its development and tracking ability are also found in more complex organisms. One of its crucial behavior is Chemotaxis - the ability to move in response to sensed concentration. This phenomenon is essential in various organisms not just to find food but for immunity, embryogenesis and many more activities. Also, it provides us with the knowledge of physiology and neural control of movements. Much work has been done previously in the field of autonomous navigation inspired by its chemotaxis network. Initial work in the field of Chemotaxis includes [1] where the author proposes a network of sensory neurons capable of successfully driving Chemotaxis via steering and controlling the angle of turn. [2] introduces motor neurons to the above network of sensory neurons to demonstrate klinokinesis. However, their network includes few mathematical calculations without neural circuits. [3] improves upon this to provide an end-to-end SNN implementation for klinokinesis. Since klinokinesis is performed, the shortest path is not guaranteed. [4] provides an adaptive klinotaxis model to ensure the shortest path of the worm. All these works help in the gradual quantitative investigation of path-planning and contour tracking of the worm. But, two major issues need to be addressed:

1. In the above-discussed works, the sensory neurons being used (ASEL and ASER) have a complex neuron model compared to simple LIF neurons. Thus its implementation on dedicated neuromorphic hardware is not possible (LOIHI processor has only LIF neurons). To get the functionality of a sensory neuron, several LIF neurons would be needed adding to the complexity of the network and the simulation time.
2. Since the worm's natural environment is 3D, the investigation should be focused on 3D medium. The implementation is based on path correction when the worm deviates from the set concentration. The worm corrects its path by aligning itself along or against the gradient by turning. In a 2D medium, a turn or deviation from the current path is sufficient to cover the entire medium due to only 2 degrees of freedom. Thus devising uniform logic for the path correction is easier. In a 3D medium, an extra degree of freedom makes it quite challenging to devise an algorithm because compared to only two directions in 2D, now we have infinite directions to choose from when path correction is needed. In other words, we need to move along a line when path correction is required in 2D, whereas we need to move across a plane when correction is required in 3D.

[5] discuss the observed motion of C. elegan in a 3D medium. The author proposes roll maneuvers, giving the organism the capability to reorient its body by combining 2D turns in the plane of dorsoventral undulations with 3D roll maneuvers enabling it to explore the whole of 3D media. Nevertheless, this work only focuses on the behavior of the worm and not on the SNN implementation. To broaden the spectrum, we also consider the qualitative behavior of Chemotaxis in other organisms. [6] discuss the sperm navigation along the helical path in 3D chemoattractant landscape.

We aim to develop a simple autonomous end-to-end solution for contour tracking in the 3D medium completely in the spiking domain. Extending the work from 2D to 3D medium is interesting because of the new degree of freedom involved without any new variable to work with, making it more challenging.

## 2 Assumption and Setting

The worm has only one sensor which senses the concentration at a particular point in space. Thus the organism has only the concentration value available to it as input. The organism already knows the set concentration, and it maneuvers accordingly by predicting the path using the current concentration value and the previous concentration value. We assume that the neural activity is much faster than the motor action, and thus in simulation, we have considered two different time scales for both. Time step-size for neural dynamics is 1000 times smaller than that of the motor action. In other words, one step of motor action has 1000 steps of neural activity. This time step-size of motor action is known as time-window, and during this period, the worm's position remains the same. This assumption is valid and of immense importance because as the hierarchy of neurons increases in the network, it is difficult to obtain perfectly matched spiking whenever needed. The neurons at the lower level of the network are bound to have some delay. It thus becomes difficult to ensure that the feedback corresponding to a neuron is taken into account for the same time instant. The time window helps us to deal with this issue. Also, the time window ensures that the spiking frequency is the same in one time window for sensory neurons making the rate encoded information transfer easier.

All the neurons used are LIF neurons. The equations are in the form such that it is compatible with the neuromorphic hardware.

$$I[i + 1] = I[i] * (1 − decay) + I\_0 * (y) \qquad (1)$$
$$V[i + 1] = V[i] * (1 − decay) + (k/C) * (I[i + 1]) \qquad (2)$$

In equation 1, $I[i]$ is the current at instant i, decay is the decay constant, $I\_0$ is some constant value. The value of y depends on the type of neuron. y can be a constant value(for constant bias current), proportional to concentration value (as in the case of sensory neurons), or a linear combination of binary spikes (post-synaptic neuron).

In equation 2, $V[i]$ is the potential built in the neuron at time instant i, decay is the decay constant, k is an arbitrary constant, and C is the capacitance.

We are considering both discrete and continuous 3D concentration space. The discrete concentration space is more in accord with the neuromorphic hardware available as of now, where the concentration has to be encoded as frequency. Since we have a time window of 1000 samples, the maximum frequency obtained is 500 (when spikes are observed every alternate time instant). Other frequencies that could be observed are 333 (when spikes are observed every 3rd time instant), 250, 200, and so on. For meaningful encoding and decoding of the information, we need a one-to-one mapping between concentration and frequency. Suppose a concentration value of 6.7 units corresponds to the frequency of 500 and 3.4 corresponds to 333, then every concentration value between 3.4 to 6.7 will propagate the same information since the frequency would be the same. This could lead to a discrepancy in the continuous concentration space simulations. Also, tracking is much better when there is some margin on both sides of the set concentration. This ensures that if we go in the opposite direction, useful information could be transferred in the network giving the worm feedback to correct its path. The discrete concentration space have the following values: 0.1, 0.2, 0.3, 0.4, 0.5, 0.6, 0.7, 0.8, 0.9, 1.0, 1.2, 1.4, 1.8, 2.3, 3.4, 6.7.

In terms of profile, we are considering Gaussian concentration profile along with linear concentration profile because key observations are easier to illustrate in linear profile in terms of contour tracking and its behavior when it reaches the set concentration.

# 3 Approach and algorithm

We only have the concentration values as rate encoded information, and we navigate the 3D space based on this information. Since we are using the path correction method, we only need to identify when the worm goes off-course. There are four control conditions as mentioned in Table: 1 which help us identify these cases.

Table 1: Control Conditions for navigation

| Setpoint Relation | Gradient Relation | Feedback |
|---|---|---|
| $C\_set > C\_current$ | $C\_current > C\_previous$ or $\Delta > 0$ | Continue moving |
| $C\_set < C\_current$ | $C\_current < C\_previous$ or $\Delta < 0$ | Continue moving |
| $C\_set > C\_current$ | $C\_current < C\_previous$ or $\Delta < 0$ | Change direction |
| $C\_set < C\_current$ | $C\_current > C\_previous$ or $\Delta > 0$ | Change direction |

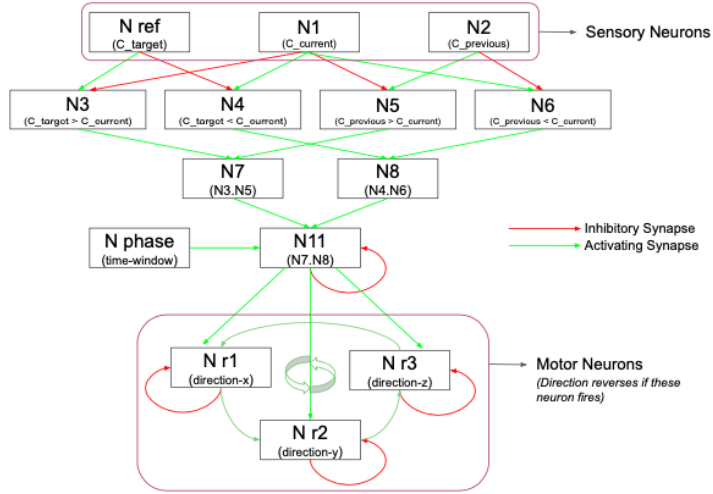

Fig. 1: Flow chart for the network of the neurons

We have few sensory neurons to sense the concentration, some motor neurons that help in the motion, and few intermediate neurons that help complete the network. Sensory neurons have current dynamics dependent on concentration values. Intermediate neurons and motor neurons have excitatory and inhibitory synaptic traces due to other neurons. We will go into more depth related to the current dynamics of the neurons in the next section.

    The neural network consists of 14 neurons. N_ref, N1, and N2 are the sensory neurons that spike corresponding to the set concentration, current concentration of the worm, and the worm's previous concentration, respectively. N2 could be considered a neuron with a long axon, and so there is a delay in propagating the information/signal giving the worm ability to sense delayed concentration or previous concentration value.

N3, N4, N5, and N6 are the intermediate neurons. Synaptic trace is a component added to the current of the post-synaptic neuron on firing of pre-synaptic neuron. A positive trace is added for activation and a negative trace for inhibition. Hence, the post-synaptic neuron will have a positive current value only if there are more positive traces than the negative traces. This implies that these intermediate neurons will spike if the spiking frequency of activating neuron is greater than the spiking frequency of inhibiting neuron. N3 spikes if the set concentration is greater than current concentration, N4 spikes if the set concentration is less than current concentration, N5 spikes if the previous concentration is greater than current concentration, and N6 spikes if the previous concentration is less than current concentration. These four intermediate neurons help identify the instances when the worm is straying from the path to the set concentration. We have to apply corrective measures if either of N7 or N8 spike.

$$N3 . N5 => (C\_set > C\_current) \text{ AND } (C\_previous > C\_current) \qquad (3)$$
$$N4 . N6 => (C\_set < C\_current) \text{ AND } (C\_previous < C\_current) \qquad (4)$$

Eq-3 and Eq-4 represent the conditions for the spiking of N7 and N8, respectively. Either of these two neurons spikes if we need to change our direction from the current path. As far as current dynamics are concerned for these two neurons, it could be thought of as the implementation of AND logic gate. That is, both these neurons spike only if both their pre-synaptic neurons spike.

N_phase is the neuron that spikes after every time window. N11 spikes if either of N7 or N8 spike indicating that path correction is required. It should be noted that N7 and N8 can often spike within a time window, but the corrective measure is required only once within a time window. It is ensured by setting a higher refractory period in N11 and by self-inhibition of N11. Multiple activations from N7 and N8 can drive up the current of N11, but we need to ensure that N11 spikes even if there is a single N7 or N8 spike. Here activation from N_phase comes into play.

We now know when we need to do the path correction. This information needs to be relayed to the motor neuron via activation from N11. All the motor neurons have self inhibitions, and they cyclically activate one neuron. This ensures that every time N11 spikes, exactly one of N_r1, N_r2 or N_r3 spike in a cyclic manner. This is how we implement cyclic directional updates. Once these neurons fire, the corresponding direction is reversed to move towards the set concentration. (eg.- If N_r1 fires, +ve x becomes -ve x and vice versa)

## 4  Current Dynamics

The whole network has 4 different types of neural activity. First, the intermediate neurons N3, N4, N5 and N6 which have similar current dynamics. Next, the AND gate implementation for neuron N7 and N8. The current dynamics for neuron N11 is different and more complex from the rest of the neurons. Finally the current dynamics for the motor neurons N_r1, N_r2, and N_r3.

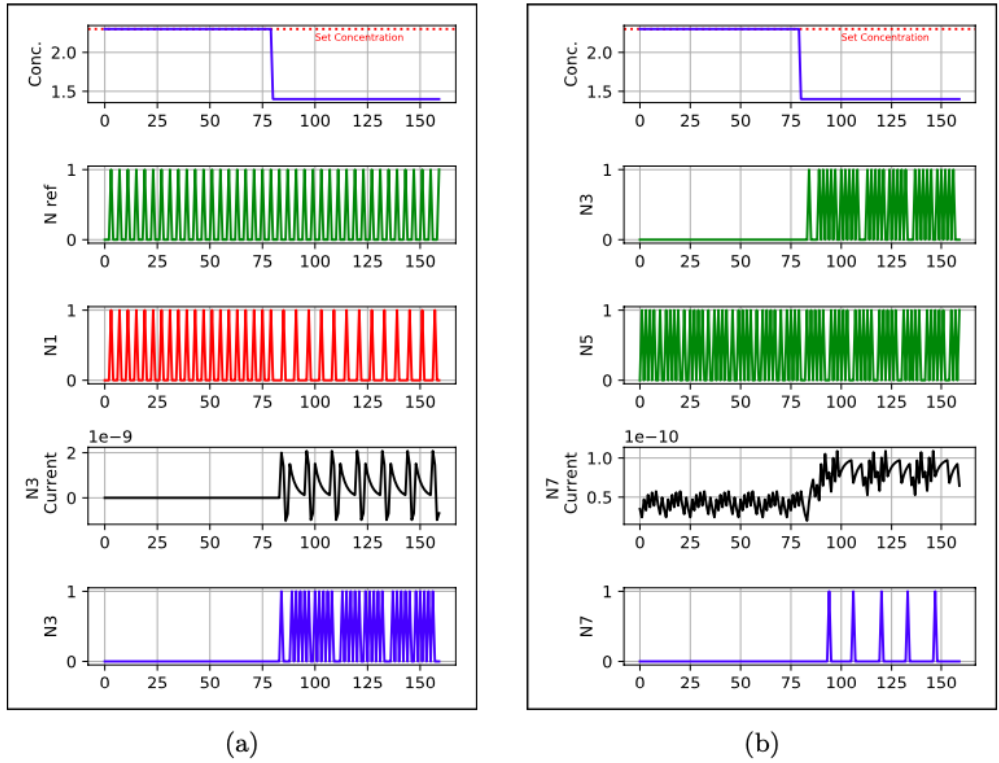

Fig. 2: Figures (a) and (b) represent the intermediate neuron's current dynamics and the AND logic gate implementation. The first row in each figure represents the sensed concentration. The set concentration is marked with a red dotted line. The following two rows represent the pre-synaptic neurons. The green color means that the neuron is activating, and the red color means that the neuron is inhibiting in nature. The fourth row represents the developed current in the post-synaptic neuron, and the last row represents the spiking pattern of the post-synaptic neuron.

In figure 2a, from the time instant 0 to 80, concentration is equal to the set concentration, and hence we see that the activating and inhibiting neuron fire with the same frequency and no current develops in N3. From time instant 80 to 160, activating neuron fires at a higher frequency and thus current develops in N3, and it spikes. The current even reaches a negative value because the pre-synaptic neurons fire at different time instants, but the activating neuron's higher spiking frequency is enough for the current to rise and the neuron to spike.

In figure 2b, from the time instant 0 to 80, the concentration is equal to the set concentration, and N3 does not spike. We know that N5 fires when C_previous is greater than C_current. Since before 0, the concentration was higher than 2.3, N5 spikes from 0 to 80 (For better clarity and understanding, we show only a tiny segment of the spiking pattern near the concentration transition). Current develops in N7 from 0 to 80, but is not sufficient for N7 to spike. From time instant 80 to 160, N3 also starts spiking because the current concentration is smaller than the set concentration. Current in N7 increases, and we observe that N7 spikes, telling the worm that it is deviating from the set concentration. Here we see that both N3 and N5 need to spike in order for N7 to spike, and hence it acts as a logical AND gate in the spiking domain.

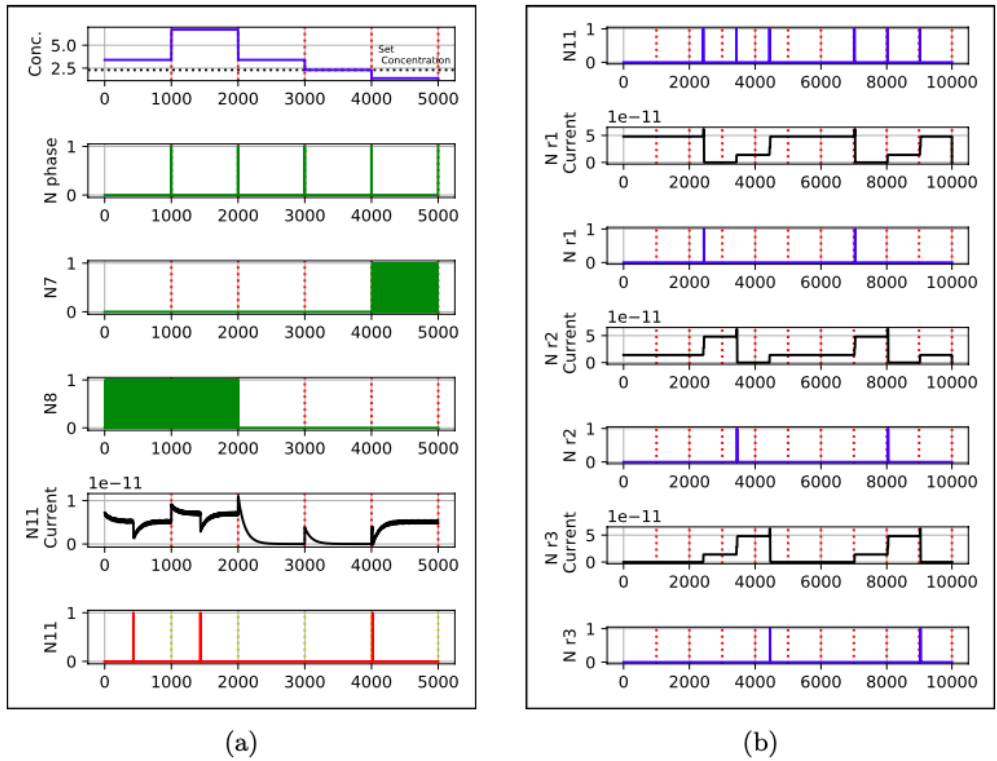

Fig. 3: Figure (a) represents the current dynamics of N11, and figure (b) represents the current dynamics for the sub-network for the cyclic directional updates. In figure (a), green spikes mean activating, and red means inhibiting neurons for the spiking of N11. (N11 is red due to self-inhibition). In figure (b), a neuron can be activating and inhibiting different neurons simultaneously. Hence they are not color-coded. For the nature of synapse, one can refer to the network shown in figure 1

Figure 3a shows that N7 and N8 spikes if the worm is moving away from the set concentration. The strength of activation from N7 and N8 is poor, and so even when they spike continuously, we observe that the current decreases because the decay rate is more prominent than the activation from N7 and N8. The activation from N phase and inhibition from N11 are both strong, and we see a sudden rise and drop in current when the respective neuron spikes.

In figure 3b, N11 acts as activating neuron for all three N_r1, N_r2, and N_r3. We also observe that once any of these neurons fire, say N_r1, its current decreases due to self-inhibition. It activates N_r2, and since N_r2 also receives activation from N11, there is a large increase in current. The next time N11 spikes, the activation is enough for N_r2 to spike, and this cycle continues.

## 5 Results

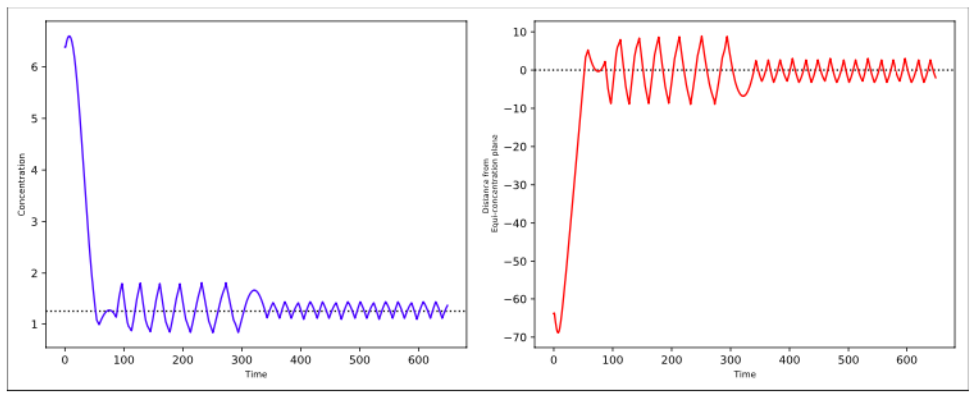

Fig. 4: Figure (a) shows the concentration variation with respect to time. Figure (b) shows the distance of worm from the equi-concentration sphere that it is supposed to track.

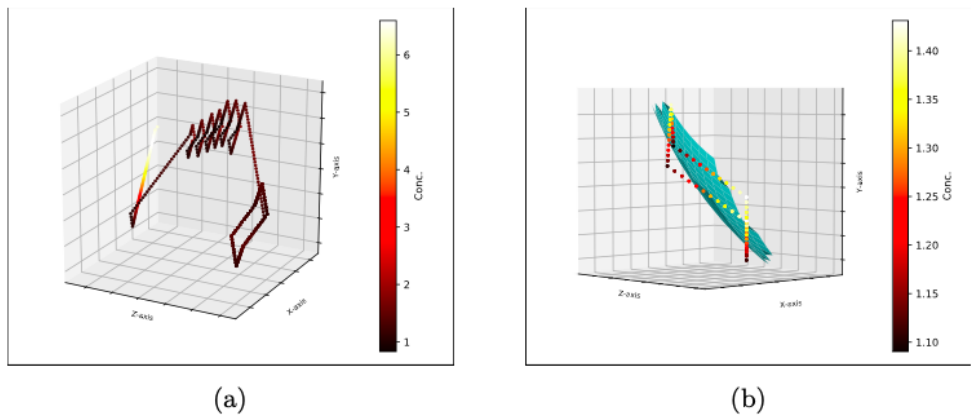

Fig. 5: Figures (a) shows the path travelled by the worm while tracking a set concentration. Figure (b) shows the behaviour of the worm once it reaches near the set concentration.

Figures 4, 5a and 5b are for Gaussian concentration space. The set concentration is 1.24 units, and the initial concentration is 6.7 units. In figure 5a, we can see that the worm starts from a white region (high concentration according to the color bar) and settles at a dark shade (low concentration). The worm reaches the set concentration initially with large oscillations where the path is similar to a distorted helix as seen in the middle section of figure 5a. Finally, the oscillation decreases when the worm attains its equilibrium position around the equi-concentration sphere. Figure 5b shows that the worm performs a closed-loop motion with deviation from the equi-concentration sphere shown by the blue surface.

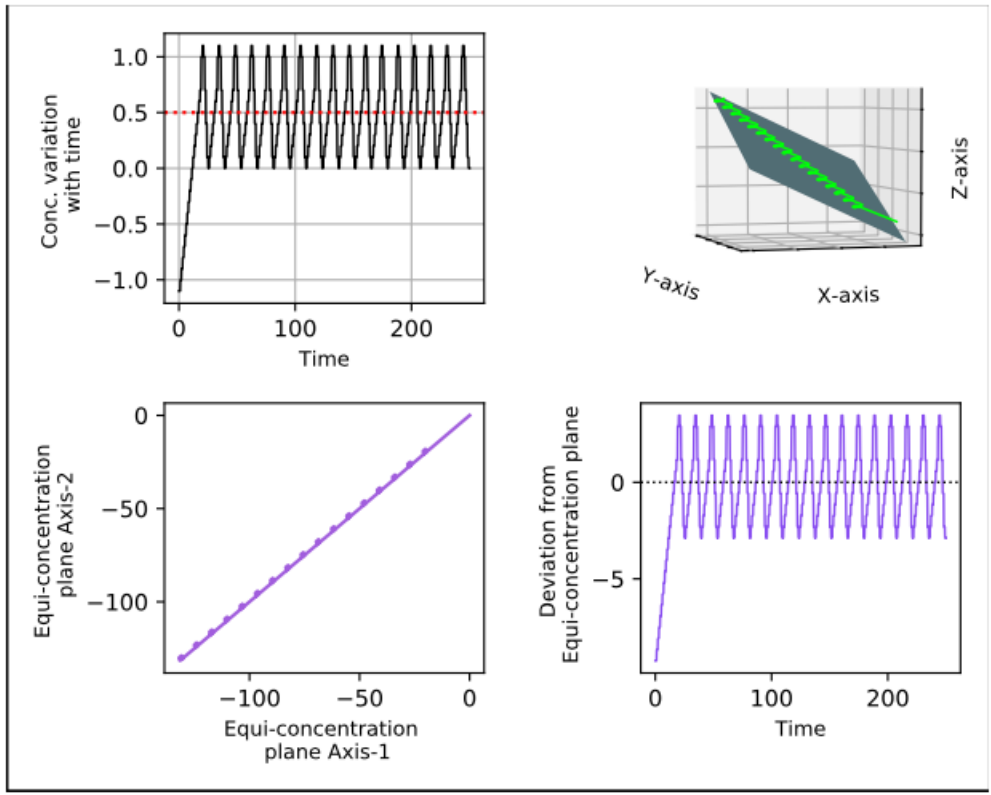

Fig. 6: Result set for linear concentration profile. Initial Concentration = -1.1 units (Negative concentration is not of any significance because N1 and N2 would not spike for these concentration values. It is negative only because of the chosen initial position and till the time it reaches a positive concentration value, the negative input cannot stimulate the sensory neurons). Set concentration = 0.5. Initial position = (14,14,15). Concentration profile: $0.6 + 0.1(x-40) + 0.1(y-20) -0.1(z-30)$. First sub-figure is for concentration variation, second sub-figure for the 3D plot of the path travelled where the gray plane represents the equi-concentration plane it is supposed to track. Third sub-figure shows the projection of motion in the equi-concentration plane. Fourth sub-figure shows the perpendicular deviation of the worm from the equi-concentration plane.

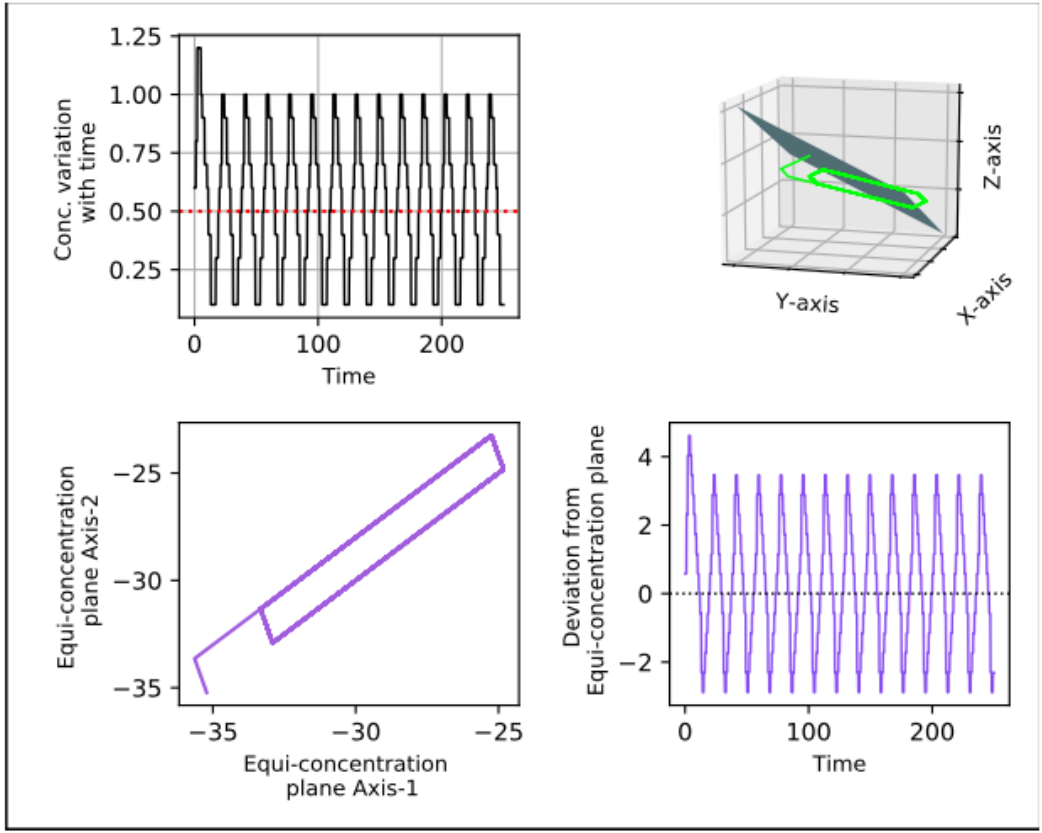

Fig. 7: Result set for discrete linear concentration profile. Initial Concentration = 0.6 units. Set concentration = 0.5. Initial position = (40,40,50). Concentration profile: 0.6 + 0.1(x-40) + 0.1(y-20) -0.1(z-30). Concentration values are discretized using the floor function with steps at the concentration values mentioned previously. The first sub-figure is for concentration variation, second sub-figure for the 3D plot of the path traversed. The gray plane represents the equi-concentration plane it is supposed to track. Third sub-figure shows the projection of motion in the equi-concentration plane. Fourth sub-figure shows the perpendicular deviation of the worm from the equi-concentration plane.

## 6 Conclusion

We have successfully devised an algorithm for spiking neural network implementation of klinokinesis for Chemotaxis, which is compatible with neuromorphic hardware. After reaching near the set concentration, the worm performs oscillatory motion around the set concentration. The motion is regular and generally closed-loop.

While tracking the concentration, the path observed is like a distorted helix, closer to the observed motion in [5]. Thus, it could be said that this simple algorithm can mimic the biologically observed motion to a significant extent.

The concentration range which provides meaningful information to the network is 0.1 to 6.7. The worm faces no issue if it tracks a value close to the median of this range, but if we start moving toward either side of the median, the deviation starts to increase. We can expand this range by using a larger time window.

As the concentration increases for continuous concentration space, the same spiking pattern is observed for a range of concentration values, and hence the propagated information is meaningless, and deviation is more.

We have successfully implemented klinokinesis keeping in mind the limitations of the dedicated neuromorphic hardware. The algorithm considers 8 possible directions to move in, from a point in space depending on the sign of the orthogonal basis directions. This is because of the difficulties that arise during the initialization of the concentration space which has to be discrete since these values have to be stored as spiking rate in the hardware.

This could be considered as the first step toward realizing much more complex algorithms to implement klinotaxis and orthotaxis in the future.

| Features | This work | [2] | [3] | [4] |
|---|---|---|---|---|
| Medium | 3D | 2D | 2D | 2D |
| Klinokinesis | ✓ | ✓ | ✓ | ✓ |
| Klinotaxis | ✗ | ✗ | ✗ | ✓ |
| Orthotaxis | ✗ | ✗ | ✗ | ✓ |
| End-to-End SNN | ✓ | ✗ | ✓ | ✓ |
| SNN Hardware-compatibility | ✓ | ✗ | ✗ | Limited |

Fig. 8: Qualitative Benchmarking